\documentclass[letterpaper, 10 pt, conference]{IEEEtran}
\IEEEoverridecommandlockouts
\usepackage{cite}
\usepackage{amsmath,amssymb,amsfonts}
\usepackage{algorithmic}
\usepackage{graphicx}
\usepackage{textcomp}
\usepackage{cite}
\usepackage{multirow}
\usepackage{hyperref}
\usepackage[font=small]{caption}
\usepackage{subcaption}
\usepackage{balance}

\usepackage[table,xcdraw]{xcolor}

\makeatletter
    \newcommand{\linebreakand}{%
      \end{@IEEEauthorhalign}
      \hfill\mbox{}\par
      \mbox{}\hfill\begin{@IEEEauthorhalign}
    }
\makeatother

\def\BibTeX{{\rm B\kern-.05em{\sc i\kern-.025em b}\kern-.08em
    T\kern-.1667em\lower.7ex\hbox{E}\kern-.125emX}}
\begin{document}


\hbadness=99999 

\title{UAV-VLA: Vision-Language-Action System for Large Scale Aerial Mission Generation
}

\author{
\IEEEauthorblockN{Oleg Sautenkov}
\IEEEauthorblockA{Skoltech\\
Moscow, Russia\\
oleg.sautenkov@skoltech.ru}\\   
\and
\IEEEauthorblockN{Yasheerah Yaqoot}
\IEEEauthorblockA{Skoltech\\
Moscow, Russia\\
yasheerah.yaqoot@skoltech.ru}\\
\and
\IEEEauthorblockN{Artem Lykov}
\IEEEauthorblockA{Skoltech\\
Moscow, Russia\\
artem.lykov@skoltech.ru}\\     
\and
\IEEEauthorblockN{Muhammad Ahsan Mustafa}
\IEEEauthorblockA{
Skoltech\\
Moscow, Russia\\
ahsan.mustafa@skoltech.ru}\\
\linebreakand
\IEEEauthorblockN{Grik Tadevosyan}
\IEEEauthorblockA{Skoltech\\
Moscow, Russia\\
grik.tadevosyan@skoltech.ru}\\
\and
\IEEEauthorblockN{Aibek Akhmetkazy}
\IEEEauthorblockA{Skoltech\\
Moscow, Russia\\
aibek.akhmetkazy@skoltech.ru}
\and
\IEEEauthorblockN{Miguel Altamirano Cabrera}
\IEEEauthorblockA{Skoltech\\
Moscow, Russia\\
m.altamirano@skoltech.ru}
\and
\IEEEauthorblockN{Mikhail Martynov}
\IEEEauthorblockA{Skoltech\\
Moscow, Russia\\
mikhail.martynov@skoltech.ru}
\linebreakand
\IEEEauthorblockN{Sausar Karaf}
\IEEEauthorblockA{Skoltech\\
Moscow, Russia\\
sausar.karaf@skoltech.ru}
\and
\IEEEauthorblockN{Dzmitry Tsetserukou}
\IEEEauthorblockA{Skoltech\\
Moscow, Russia\\
d.tsetserukou@skoltech.ru}
}



\maketitle

\begin{abstract}
The UAV-VLA (Visual-Language-Action) system is a tool designed to facilitate communication with aerial robots. 
By integrating satellite imagery processing with the Visual Language Model (VLM) and the powerful capabilities of GPT, UAV-VLA enables users to generate general flight paths-and-action plans through simple text requests. 
This system leverages the rich contextual information provided by satellite images, allowing for enhanced decision-making and mission planning. 
The combination of visual analysis by VLM and natural language processing by GPT can provide the user with the path-and-action set, making aerial operations more efficient and accessible. The newly developed method showed the difference in the length of the created trajectory in 22\% and the mean error in finding the objects of interest on a map in 34.22 m by Euclidean distance in the K-Nearest Neighbors (KNN) approach. Additionally, the UAV-VLA system generates all flight plans in just 5 minutes and 24 seconds, making it 6.5 times faster than an experienced human operator. The code is available here: 
\href{https://github.com/sautenich/uav_vla}{https://github.com/sautenich/uav-vla}

\end{abstract}


\section{Introduction}

\begin{figure}[t!]
\centering
\includegraphics[width=0.81\linewidth]{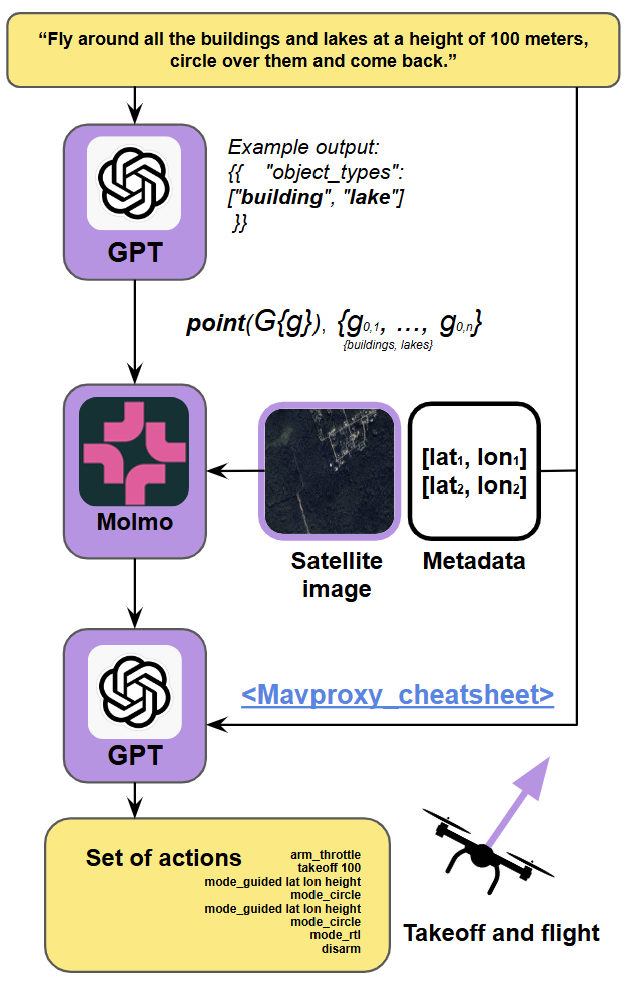} 
\caption{The pipeline of the UAV-VLA system.}
\label{software}
\end{figure}

In recent years, the field of aerial robotics has witnessed significant advancements, particularly in the development of unmanned aerial vehicles (UAVs) and their applications across various domains such as surveillance, agriculture, and disaster management\cite{ai_meets_uav}. As the complexity of missions increases, the need for effective communication between human operators and UAVs has become crucial. Traditional methods of interaction often rely on complex programming or manual controls, which can be cumbersome and limit the accessibility of these technologies to a broader audience. Previous work in this domain has explored various approaches to enhance the traditional human-UAV communication. Most of the systems mainly focused on manual piloting and basic waypoint navigation\cite{sautenkov2024flightararflightassistance}, which required extensive training and experience. 


More recent advancements have introduced automation and semi-autonomous UAV systems, allowing for improved mission planning and execution. Transformer-based models \cite{vaswani2023attentionneed} can generate outputs that represent actions for a robot. For instance, they can produce a set of positions for a robotic gripper in systems like OpenVLA and RT \cite{kim2024openvlaopensourcevisionlanguageactionmodel}, \cite{brohan2023rt1roboticstransformerrealworld}, \cite{brohan2023rt2visionlanguageactionmodelstransfer}.  The works\cite{lee2024citynavlanguagegoalaerialnavigation}, \cite{zhong2023safervisionbasedautonomousplanning}, \cite{fan2023aerialvisionanddialognavigation}, \cite{EmbodiedCityzhang2024}  generates a sequence of movements as output, referred to as Visual Language Navigation (VLN) models.

Many VLA and VLN approaches require large datasets with paired language instructions and agent behavior, but struggle to generalize to new environments and lack a global-scale understanding. Our research focuses on developing systems that generate path plans and execute actions based solely on linguistic instructions and open satellite data, leveraging zero-shot capabilities of powerful models without additional training.

Our contributions are as follows:
\begin{itemize}
    \item We present a large-scale \textbf{Vision-Language-Action (VLA)} system that generates path-action sets from text-based mission requests using satellite images.
    \item We introduce the \textbf{UAV-VLPA-nano-30} benchmark for evaluating VLA systems at a global scale.
    \item We validate our system on UAV-VLPA-nano-30, showing performance comparable to human-level path and action generation.
\end{itemize}


\section{Related Work}


The introduction of Vision Transformers (ViT) \cite{visiontransformerdosovitskiy2021imageworth16x16words},\cite{CLIPradford2021learningtransferablevisualmodels} marked a significant advancement in the development of full-fledged models capable of processing and integrating multiple types of input and output, including text, images, video, and more. Building on this progress, OpenAI introduced models like ChatGPT-4 Omni\cite{openai2024gpt4technicalreport}, which can reason across audio, vision, and text in real time, enabling seamless multimodal interactions. To address the problem of objects finding in robotics applications, Allen Institute of AI introduced model Molmo, that can point the requested objects on an image\cite{deitke2024molmopixmoopenweights}.

The usage of the transformer-based models allowed the extensive developing of the new methods, benchmarks, and datasets for Vision Language Navigation tasks. Firstly, the problem of Aerial Visual Language Navigation was proposed by Liu et al. \cite{liu2023aerialvlnvisionandlanguagenavigationuavs}, where they introduced the Aerial VLN method together with AerialVLN dataset. In \cite{fan2023aerialvisionanddialognavigation} Fan et al. described the simulator and VLDN system, that can support the dialog with an operator during the flight.
Lee et al.\cite{lee2024citynavlanguagegoalaerialnavigation} presented an extended dataset with geographical meta information (streets, squares, boulevards, etc.). The introduction of dataset was paired with the new approach for goal predictor. Zhang et al. 
\cite{EmbodiedCityzhang2024} took a pioneering step by building a universal environment for embodied intelligence in an open city. The agents there can perform both VLA and VLN tasks together online. Gao et al.
\cite{gao2024aerialvisionandlanguagenavigationsemantictopometric} presented a method, where a map was provided as a matrix to the LLM model. In that work was introduced the Semantic Topo Metric Representation (STMR) approach, that allowed to feed the matrix map representation into the Large Language Model. In \cite{wang2024realisticuavvisionlanguagenavigation} Wang et al. presented the benchmark and simulator dubbed OpenUAV platform, which provides realistic environments, flight simulation, and comprehensive algorithmic support.


Google DeepMind introduced the RT-1 model in their study \cite{brohan2023rt1roboticstransformerrealworld}, wherein the model generates commands for robot operation. The researchers collected an extensive and diverse dataset over several months to train the model. Utilizing this dataset, they developed a transformer-based architecture capable of producing 11-dimensional actions within a discrete action space. Building on the foundation of RT-1, the subsequent RT-2 model \cite{brohan2023rt2visionlanguageactionmodelstransfer} integrates the RT-1 framework with a Visual-Language Model, thereby enabling more advanced multimodal action generation in robotic systems. The work of \cite{bivla} and \cite{berman2024missiongptmissionplannermobile} highlights the potential of transformers and end-to-end neural networks to handle complex vision-language-action (VLA) tasks in real time.

\section{Data and Benchmark} 
\label{benchmark}

\subsection{Satellite Images And Metadata description}
To estimate the overall success of the proposed system, we introduce a novel benchmark dataset \textbf{UAV-VLPA-nano-30} to evaluate the effectiveness of the UAV-VLA. This benchmark comprises 30 high-resolution satellite images collected from the open-source platform \href{https://earthexplorer.usgs.gov/}{USGS EarthExplorer}. Designed specifically for mission generation in aerial vehicles, the benchmark provides a standardized testbed to assess the UAV-VLA system's ability to interpret linguistic instructions and generate actionable navigation plans.

The benchmark spans diverse locations across the \textbf{United States}, including urban, suburban, rural, and natural environments. These include: buildings (living houses, warehouses), sport stadiums, water bodies (ponds, lakes), transportation infrastructure (crossroads, bridges, roundabouts), fields, and parking lots. The benchmark satellite images were captured during the \textbf{spring and summer seasons} under \textbf{daytime conditions}, ensuring clear visibility and consistent lighting.

\begin{figure}[h] 
\includegraphics[width=0.999\linewidth]{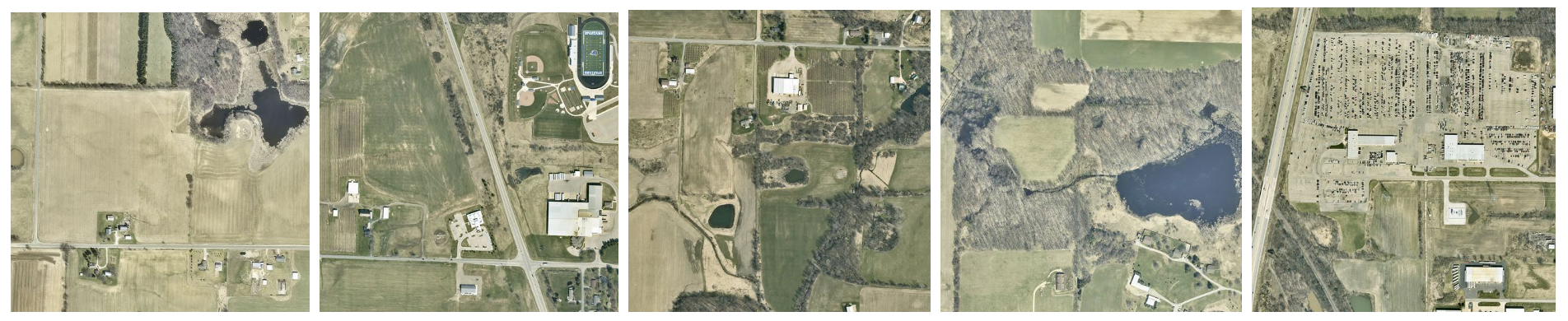}
\captionof{figure}{Examples of the satellite imagery in the benchmark data.}
\label{Fig: Dataset}
\end{figure}

The satellite imagery has a resolution of approximately \textbf{1.5 meters per pixel}, providing detailed visual representation of natural and man-made features. Each image spans an area of roughly \textbf{760 sq. meters}, offering sufficient geographic coverage for mission generation tasks. Each of the image has a metadata (geographic location description), allowing the calculation of the identified points in latitude and longitude for flight plan generation.


The dataset’s reliance on real-world satellite imagery ensures authentic representation of the environments and scenarios UAVs encounter in practical applications. 

\subsection{Manual Flight Plan Generation}
\label{manual-flight-plan}
An experienced drone operator was tasked with generating flight plans for the benchmark images. For each image, the operator was instructed: \textit{“Create a flight plan for a quadcopter to fly over all buildings inside the violet square. Height is not considered”.} The violet square boundaries were defined in Mission Planner using image metadata, and the home position was set at 10\% of the width and height from the top left corner.

\begin{figure}[h]
\centering
\includegraphics[width=0.9\linewidth]{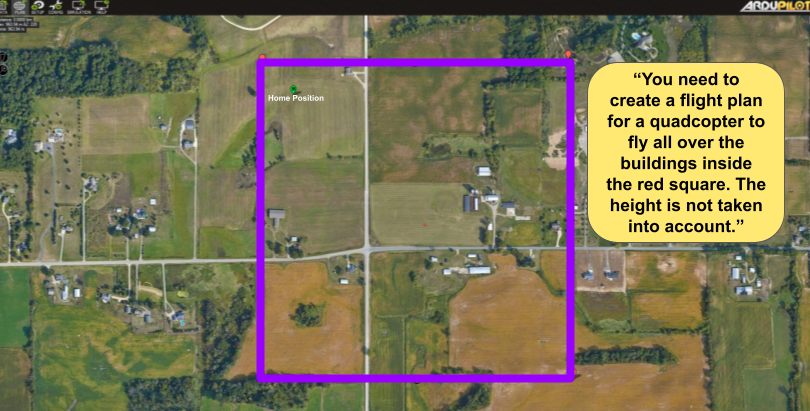}
\captionof{figure}{Mission Planner environment showing the violet square boundary, home position and buildings with the text description.}
\label{fig:mp_environment}
\end{figure}

The operator manually created flight plans for all 30 images in 35 minutes. An example of the Mission Planner environment with the violet square and home position is shown in Fig.~\ref{fig:mp_environment}.
The total length of the UAV-VLPA-nano-30 benchmark samples was 63.89 km and the average length was 2.13 km created by the operator using the Mission Planner.

\section{Methodology}
\label{methodology}
In this paper, we present a novel UAV-VLA system that leverages Large Language Models (LLMs) and Vision Language Models (VLMs) for action prediction in aerial tasks. As shown in Fig.~\ref{software}, the framework comprises three key modules: \textbf{goal extracting GPT module, object search VLM module, and actions generation GPT module.}

The process begins with a language instruction:
\begin{equation}
\ I = \bigl\{i_1, i_2, \dots, i_k\bigl\},
\label{eq:instruction}
\end{equation}
where $I$ is the input prompt of length $k$, varying by task complexity. For instance: “Fly around all the buildings at a height of 100 meters and come back.”

The \textbf{goal extracting GPT module} parses $I$ into a set of goals:
\begin{equation}
\ G = GPT(I) = \bigl\{g_1, g_2, \dots, g_n\bigl\},
\label{eq:goal}
\end{equation}
where $G$ contains goals derived from the instruction, tailored to the task.

The \textbf{object search VLM module} identifies these goals in the satellite image, producing processed points:
\begin{equation}
\ P_p = Molmo(G) = \bigl\{[g_{1,1}, g_{1,2}], [g_{2,1}, g_{2,2}], \dots, [g_{n,1}, g_{n,2}]\bigl\}
\label{eq:processed_points}
\end{equation}

These points are transformed into global coordinates using metadata:
\begin{equation}
\ P_g = f(P_p) = \bigl\{[lat_{1,1}, lon_{1,2}], \dots, [lat_{n,1}, lon_{n,2}]\bigl\},
\label{eq:global_points}
\end{equation}
ensuring accurate mapping to real-world locations.

Finally, the \textbf{actions generation GPT module} uses $P_g$, mission details, and MAVProxy \cite{mavproxy} to generate UAV actions:
\begin{equation}
\ A = GPT(P_g, [A_b]) = \bigl\{A_1, A_2, \dots, A_n\bigl\}
\label{eq:actions}
\end{equation}

This pipeline integrates instruction parsing, object detection, and coordinate transformation, enabling UAVs to autonomously generate precise mission plans tailored to specific tasks.

\section{Experiments}
This section evaluates the UAV-VLA system using the benchmark introduced in Section \ref{benchmark}, focusing on flight plan creation and a novel evaluation metric to assess system effectiveness.

\subsection{Evaluation Metrics}
The evaluation metric considers two aspects: the total \textbf{length} of the path and the \textbf{error} between system-generated and human-generated (ground truth) trajectories.

Three methods were used to compute the error: The \textbf{Sequential Method} aligns points step-by-step but is prone to cumulative errors. \textbf{Dynamic Time Warping (DTW)} \cite{DTW2007} allows non-linear alignment to measure path similarity. \textbf{K-Nearest Neighbors (KNN)} matches system points to the nearest ground-truth points, offering a general accuracy measure without considering order.





The error is quantified using the Root Mean Square Error (RMSE) calculated using Eq.~\ref{eq:rmse}:
\begin{equation}
\
RMSE = \sqrt{\frac{1}{n} \sum_{i=1}^{n} {(x_n - \hat{x}_n)^2 + (y_n - \hat{y}_n)^2}},
\label{eq:rmse}
\end{equation}
where $x_n$ and $\hat{x}_n$ are the system-generated and ground-truth points, respectively, and $n$ is the total number of points.

\subsection{System Setup and Procedure}
The system was evaluated using the command described in Section \ref{methodology}: \textit{“Create a flight plan for the quadcopter to fly around each building at a height of 100 m, return to home, and land at the take-off point”.} The experiment was conducted on a PC with an RTX 4090 graphics card (24GB VRAM) and Intel Core i9-13900K processor. Due to memory constraints, the quantized Molmo-7B-D BnB 4-bit model \cite{molmo_quantized} was used.

\begin{figure}[h]
\centering
\begin{subfigure}{.24\textwidth}
  \centering
  \includegraphics[width=.9\linewidth]{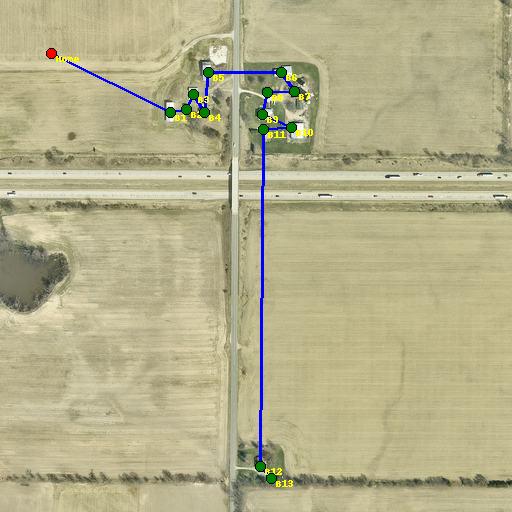}
  \caption{Human-made}
  \label{fig:sub1}
\end{subfigure}%
\begin{subfigure}{.24\textwidth}
  \centering
  \includegraphics[width=.9\linewidth]{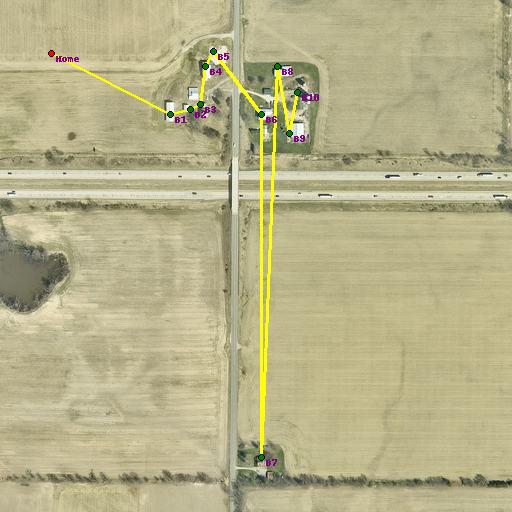}
  \caption{UAV-VLA}
  \label{fig:sub2}
\end{subfigure}
\caption{Comparison of flight plans generated by a human expert (a) and the UAV-VLA system (b).}
\label{fig:test}
\end{figure}

We compared flight plans generated by the UAV-VLA system with human-generated plans. Fig.~\ref{fig:test} shows an example comparison.

\section{Experimental Results}
\begin{figure}[h]
    \centering
    \includegraphics[width=1.0\linewidth]{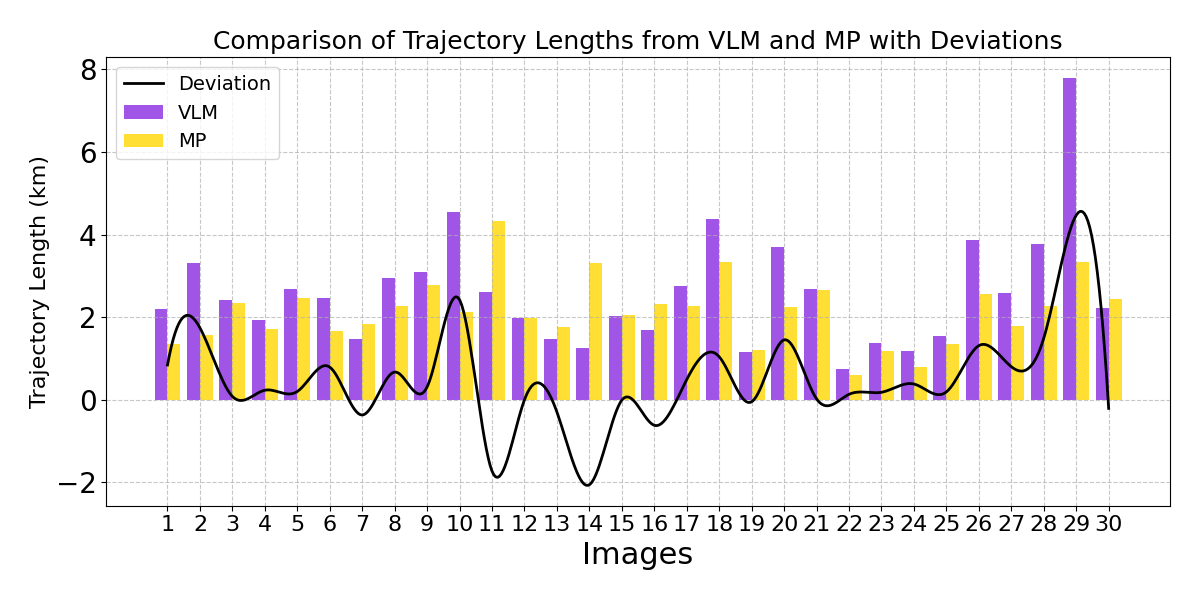}
    \caption{The comparison of the trajectory lengths made by UAV-VLA and by an experienced operator.}
    \label{fig:traj_length}
\end{figure}

The newly developed system has shown a general trajectory length of 77.74 km on the benchmark UAV-VLA-nano-30, which is 13.85 km, or 21.6\%, longer than the ground-truth trajectory created by the experienced UAV pilot (Sec.~\ref{manual-flight-plan}). In 7 out of 30 cases, or 23\% of the cases, the UAV-VLA generated a trajectory path that was even shorter as can be seen in Fig.~\ref{fig:traj_length}.

\begin{figure}[h]
    \centering
    \includegraphics[width=0.9\linewidth]{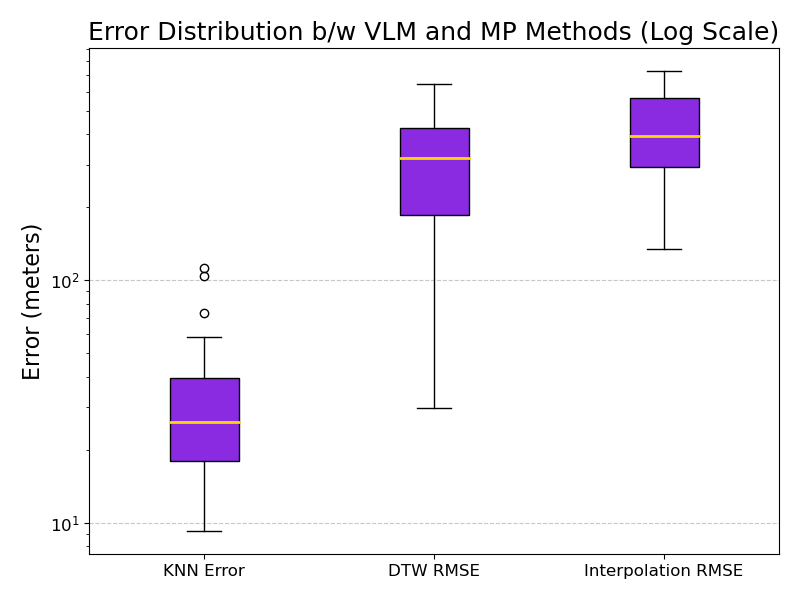}
    \caption{The error of UAV-VLA in comparison with ground truth.}
    \label{fig:errors}
\end{figure}

As shown in Table~\ref{table:errors}, the sequential RMSE exhibited the largest mean error of 409.54 m per trajectory, which was expected due to its strict reliance on the sequential order of points. The Dynamic Time Warping (DTW) method demonstrated a reduced mean error of 307.27 m, highlighting its ability to account for temporal variations more effectively. The K-Nearest Neighbors (KNN) method resulted in the smallest mean error, as it disregards the sequence entirely and focuses solely on the spatial proximity of points.

\begin{table}[htbp]
\caption{\textsc{Comparison of RMSE Metrics for Different Methods}}
\begin{center}
\begin{tabular}{|c|c|c|c|}
\hline
\rowcolor[HTML]{FFE972} 
\textbf{Metric (RMSE)} & \textbf{KNN (m)} & \textbf{DTW (m)} & \textbf{Sequential (m)} \\
\hline
Mean   & 34.22   & 307.27   & 409.54   \\
\hline
Median & 26.05   & 318.46   & 395.59   \\
\hline
Max    & 112.49  & 644.57   & 727.94   \\
\hline
\end{tabular}
\label{table:errors}
\end{center}
\end{table}

The UAV-VLA system processes all benchmark images in approximately 5 minutes 24 seconds, 2 minutes for identifying required points using the object search VLM module and 3 minutes and 24 seconds for generating mission files with the actions generation GPT module. This is 6.5 times faster than the human-generated flight plans mentioned in Sec.~\ref{manual-flight-plan}.

\section{Conclusion}
This paper presents a global-scale UAV mission generation approach, enhancing flexibility and accuracy in mission planning. By overcoming the limitations of traditional manual methods, this approach is highly beneficial in scenarios where human intervention is inefficient. Key contributions include:

\begin{itemize}
    \item The UAV-VLPA-nano-30 benchmark, providing a standardized framework for evaluating and comparing global path planning techniques.
    \item A method for interpreting natural language requests into flight paths, generating paths that are 21.6\% longer but 6.5 times faster than human-created ones, demonstrating significant efficiency.
    \item A new task for UAVs: language-based path planning, enabling UAVs to autonomously execute mission plans from natural language inputs.
\end{itemize}

This method simplifies human-UAV interaction by enabling direct natural language communication and lays the foundation for seamless robot-robot collaboration in dynamic environments.




\section{Future Work}
Future work will focus on creating a specialized dataset for satellite map-based path planning to improve model precision and efficiency in UAV mission generation. Additionally, we aim to develop an end-to-end model that autonomously generates mission plans from high-level goals, integrating action generation, path planning, and decision-making for fully autonomous UAV mission planning across diverse environments.


\section*{Acknowledgements} 
Research reported in this publication was financially supported by the RSF grant No. 24-41-02039.

\clearpage
\balance

\bibliographystyle{IEEEtran}

\bibliography{conference_101719}

\end{document}